\documentclass{article}
\usepackage{spconf,amsmath,graphicx}
\usepackage{amssymb}
\usepackage{subfigure}
\usepackage{float}
\usepackage{hyperref}
\usepackage{multirow}
\usepackage{flushend}

\title{Learning to navigate image manifolds induced by generative adversarial networks for unsupervised video generation}
\name{Isabela Albuquerque$^{1,*}$, Jo\~ao Monteiro $^{1,2,*}$, Tiago H. Falk$^1$}
\address{$^1$Institut national de la recherche scientifique (INRS-EMT), Quebec, Canada \\ $^2$Centre de Recherche Informatique de Montr\'eal (CRIM), Quebec, Canada \\
$^*$Equal contribution \thanks{Correspondence to \texttt{isabelamcalbuquerque@gmail.com}, \texttt{joaomonteirof@gmail.com}}}
\begin{document}
%
\maketitle
\begin{abstract}

In this work, we introduce a two-step framework for generative modeling of temporal data. Specifically, the generative adversarial networks (GANs) setting is employed to generate synthetic scenes of moving objects. To do so, we propose a two-step training scheme within which: a generator of static frames is trained first. Afterwards, a recurrent model is trained with the goal of providing a sequence of inputs to the previously trained frames generator, thus yielding scenes which look natural. The adversarial setting is employed in both training steps. However, with the aim of avoiding known training instabilities in GANs, a multiple discriminator approach is used to train both models. Results in the studied video dataset indicate that, by employing such an approach, the recurrent part is able to learn how to coherently navigate the image manifold induced by the frames generator, thus yielding more natural-looking scenes.

\end{abstract}
%
%
\section{Introduction}

Generative adversarial networks (GANs) \cite{goodfellow2014generative} were recently introduced as an unsupervised approach to generative modeling, employing game-theoretic training schemes in order to learn a given probability density, implicitly defined by training data. Under this setting, two models are trained jointly. The generator tries to map low dimensional samples from some simple prior distribution to higher-dimensional structured data, while the discriminator, on the other hand, tries to determine whether samples are genuine or generator outputs. To date, state-of-the-art results have been obtained for GAN-based generative modeling of images \cite{radford2015unsupervised, karras2017progressive} and audio, if image-like spectrogram representations are used \cite{donahue2018synthesizing, cai2018attacking}. However, their applications in other domains, such as temporal or discrete data, remain open problems under active investigation. Here, we direct our focus to adversarially learned video modeling.

A common strategy in recent attempts on training GANs for natural scenes generation focuses on splitting the task into simpler parts. In \cite{vondrick2016generating}, for instance, there are independent modules for foreground and background modeling. In both \cite{saito2017temporal} and \cite{tulyakov2017mocogan}, motion and frame content are learned by different parts of the architecture designed specifically for each of those aspects. In turn, in \cite{ohnishi2017hierarchical} authors tackle the problem by conditioning generation on optical flows provided \emph{a priori}. However, in all such cases, even though the model architectures are designed aiming to focus on different aspects of video generation, training is performed together, which might yield relevant training difficulties such as mode collapse and divergence \cite{goodfellow2016nips}. This is due to the higher dimensionality of videos which also include a temporal component.

In this work, we further exploit the idea of splitting the video generation process into smaller and simpler components. Frame content and motion modeling are achieved by independent blocks within the complete model: a convolutional block responsible to map a low-dimensional vector into a frame, and a recurrent block intended to receive a fixed-dimension vector input and to output a sequence of vectors to be used as inputs into the convolutional block, thus yielding a sequence of frames. Moreover, a new training scheme is devised on top of the proposed setting to avoid common issues faced when training GANs. Each of the above mentioned components, i.e. the generative model of frames as well as the generative model of sequences of frames, are trained separately. More specifically, the multi-discriminator setting introduced in \cite{neyshabur2017stabilizing} is used in both steps to further stabilize training and produce diverse generators.


Under the described setting, the frames generator can be seen as a parametric representation of the manifold of video frames, i.e. a mapping from a much lower-dimensional space to actual frames. This model is trained first and, as we obtain good samples in terms of quality and diversity, the sequence component is trained. The sequence generator, in turn, is a recurrent model trained with the goal of learning how to effectively traverse the manifold induced by the pre-trained frames generator in such a way that yields coherent frame sequences.

The remainder of this paper is organized as follows. In Section \ref{sec:gans} we briefly review Generative Adversarial Networks training. In Section \ref{sec:approach} we describe the proposed approach and provide experiments to validate it in Section \ref{sec:exper}. We provide conclusions and future research directions in Section \ref{sec:conc}.   

\section{Generative Adversarial Networks}\label{sec:gans}
GANs are generally composed of a discriminator model $D(x):\mathbb{R}^n \rightarrow \left[0,1\right]$, where $n$ is the dimensionality of the input space, and a generator $G(z):\mathbb{R}^m \rightarrow \mathbb{R}^n$, where $m$ is the size of an input noise vector $z$. $D(x)$ receives a sample from the data distribution $p_{data}$ or a sample from the generator $G(z)$, $z \sim p_{z}$. During training, its goal is to learn how to tell apart these two different types of inputs. The generator, on the other hand, aims at \textit{fooling} the discriminator by learning how to produce samples as close to the data distribution as possible. GAN training was originally defined as a min-max game, but here we utilize the non-saturating game, as defined in \cite{goodfellow2016nips}. According to this training scheme, the discriminator loss $\mathcal{L}_{D}$ and the generator loss $\mathcal{L}_G$ are respectively defined as
\begin{equation}\label{eq:disc}
\mathcal{L}_{D} = -\mathbb{E}_{x \sim p_{\text{data}}} \log D(x) -\mathbb{E}_{z \sim p_{z}} \log (1 - D(G(z))),
\end{equation}
\begin{equation}\label{eq:gen}
\mathcal{L}_G = -\mathbb{E}_{z \sim p_{z}} \log D(G(z)).
\end{equation}

With the success of GANs and its popularization, deeper analyses have shown that these models may suffer from instability during training \cite{berthelot2017began, arjovsky2017wasserstein} which can lead to lack of diversity and poor quality on the generated samples. In order to alleviate these issues, many GAN variations were proposed in the last few years \cite{neyshabur2017stabilizing, lin2017pacgan, yadav2018stabilizing}. One interesting approach proposed in \cite{neyshabur2017stabilizing} consists in using multiple discriminators where each one considers as input a low-dimensional randomly projected version of the original input. The authors empirically showed that this method yielded more stable training and provided more diverse and better quality generated samples.    

\section{Proposed Model and Training} \label{sec:approach}

The proposed method relies on two main components: (i) a convolutional frame generator $G_F$, and (ii) a recurrent model for generating videos $G_V$. The goal is to disentangle image quality and temporal coherence components of a video and letting each of the generative models individually focus on one of these two aspects. By doing so, the performance of the model relies on the capability of the frame generator to provide good and diverse images as well as on the sequence generator to be able to sequentially sample frames (i.e. navigate through the frames manifold induced by $G_F$) in a coherent order.

One of the main challenges in such an approach is to be able to train $G_F$ with enough diversity. Several methods have been proposed recently targeting mode dropping in the GAN setting \cite{lin2017pacgan}. In our experiments, we found the multiple-discriminators approach introduced in \cite{neyshabur2017stabilizing} to yield better stability during training, as well as higher sample quality and diversity. Training follows the usual steps, i.e. each discriminator is separately updated, but when updating the generator parameters, the average of discriminator losses is considered. Thus, instead of using (\ref{eq:gen}) as the generator loss during training, we use (\ref{eq:gen2}) instead, namely:

\begin{equation}\label{eq:gen2}
\mathcal{L}_G = -\sum_{k=1}^K \mathbb{E}_{z \sim p_{z}} \log D_k(G(z)), 
\end{equation}
where $D_k((G(z))$ indicates the output of the $k$-th discriminator and $K$ the total number of discriminators.  
Training of $G_F$ was performed with $K=48$ discriminators. An architecture similar to DCGAN \cite{radford2015unsupervised} was employed.

\begin{figure}
\centering
\includegraphics[width=0.43\textwidth]{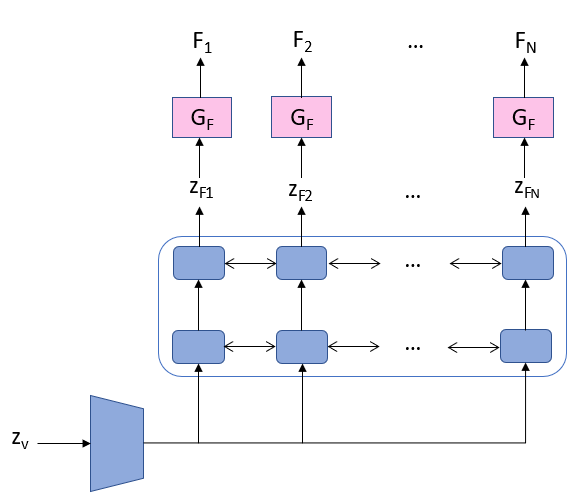}
\caption{Graphical representation of the video generator. Pink blocks represent the pre-trained frame generator.}
\label{fig:model}
\end{figure}

$G_V$ is composed of three main building blocks: an encoding stack of dense layers responsible to map a noise vector $z_v$ into a sequence of high-dimensional vectors. This sequence is fed into a bi-directional recurrent block that computes a sequence of temporally dependent $z_{Fi}$ noise vectors which are then used to sample from $G_F$. Finally, for the case of videos with length $N$, the output is obtained by sampling $N$ times from the frames generator and ordering the samples to form the final sequence $F=(F_1, \ldots, F_N)$. The described framework is represented in Fig.~\ref{fig:model}. The encoder (trapezoid) is parametrized by fully-connected (FC) layers, and the recurrent model by a two-layer bi-directional LSTM.

The scheme proposed in \cite{neyshabur2017stabilizing} was also used to train the sequence generator. In this case, we utilized 16 discriminators which inputs are reduced-dimension random projections of each frame composing the video input. It is important to highlight that $G_F$ parameters are kept unchanged during the training of $G_V$. The architectures used for the video generation GAN were: 1) Generator: FC$[100\times 512\times1024\times2048\times3840]  \rightarrow \textnormal{Bi-LSTM} [30\times128, 30\times256] \rightarrow \textnormal{FC}[512, 100]$; 2) Discriminator: similar to \cite{radford2015unsupervised} but with 3D convolutions in the place of 2D in order to take into account the temporal dimension. Random projections were implemented as norm $1$ convolutions.

\section{Experimental Results} \label{sec:exper}
In order to evaluate the proposed method, we performed two main experiments. First, we aim to show that our approach is able to generate videos with both frame quality and temporal coherence. For that, we train the frame generator using frames from the same videos used from training the videos generator. 
Overall, our goal is to investigate what the video generator is learning in terms of navigation throughout the implicit manifold parametrized by $G_F$. To this end, we plot the 2-dimensional isomap \cite{tenenbaum2000global} of generated latent variables $z_{Fi}$ by the video generator resulting from the first experiment.    
We thus built a training dataset composed of 100,000 samples from bouncing balls data \cite{srivastava2015unsupervised}. Each example consists of 30 frames-long videos with three balls bouncing. Randomly sampled frames from the same set of videos were used to train the frames generator in advance. RMSprop optimizer with learning rate equal to $0.0002$ and $0.0003$ was employed to train $G_V$ and $G_F$, respectively. $G_F$ was trained for 50 epochs with mini-batches of size 64, while 15 epochs were used for $G_V$ with mini-batches of size 8. Random seed was previously set to 10 before all experiments. A single NVIDIA GTX 1080Ti was used for training. A Pytorch \cite{pytorch} implementation is available at Github \cite{framegan}.

\subsection{Generating videos} 
\begin{figure}
\centering
\includegraphics[width=0.35\textwidth]{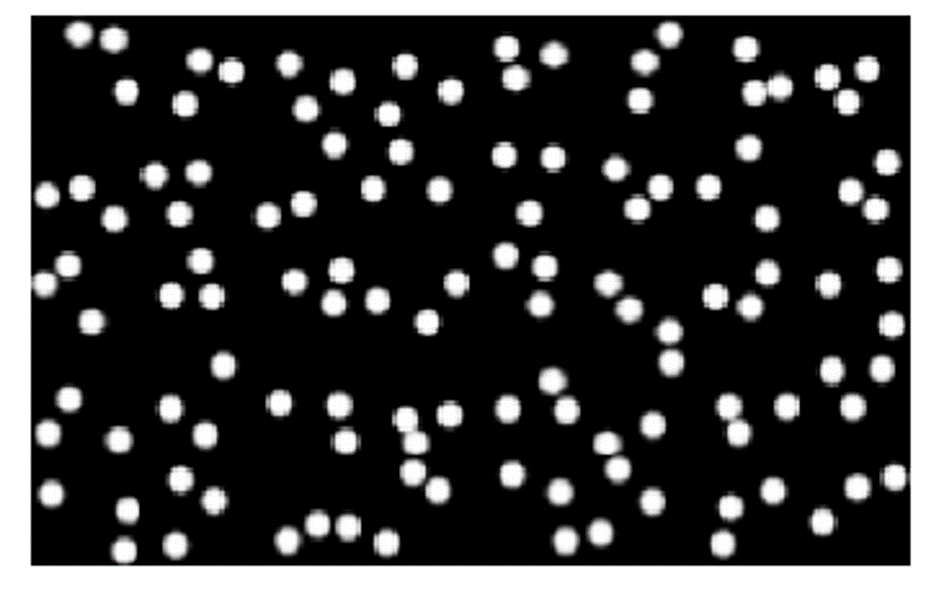}
\caption{Random samples from the frame generator.}
\label{fig:frame_gen}
\end{figure}

\begin{figure*}
\center
\subfigure[fig:real][Samples from the training data.]{\includegraphics[width=0.95\textwidth, height = 1.5cm]{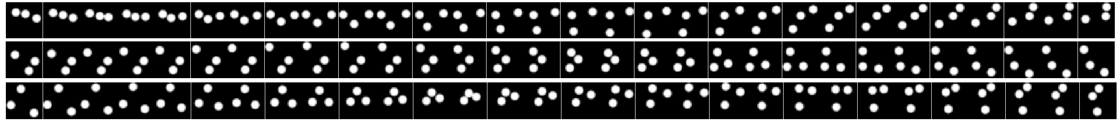}}
\qquad
\subfigure[fig:gen][Samples of videos generated by the proposed approach.]{\includegraphics[width=0.95\textwidth, height = 2cm]{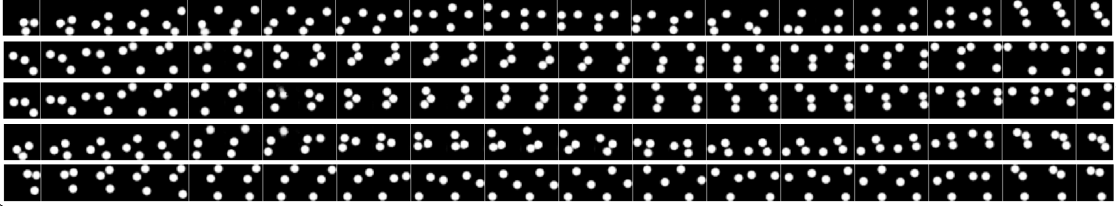}}
\caption{Real data (a) and generated video samples (b). $G_F$ and $G_V$ were both trained using the three bouncing balls dataset. Time increases from left to right. }
\label{fig:results}
\end{figure*}

In Fig.~\ref{fig:frame_gen}, we show samples randomly drawn from the frame generator. By visual inspection, we notice that, as desired, good quality and diversity were obtained. Using this model as $G_F$, we train $G_V$ and show random samples in Fig.~\ref{fig:results}b. To provide a reference for comparison, we also show in Fig.~\ref{fig:results}a three randomly selected video samples drawn from the real data distribution. Each frame is plotted individually such that time increases from left to right. Visual inspection of generated sequences of frames indicates that both the quality of individual frames (as ensured by the frame generator) and temporal coherence were close to original samples. More specifically, we notice that most of the transitions between frames are as smooth as in the original data samples. 

We further highlight that the generated video samples are diverse, which suggests the proposed training scheme is effective in avoiding strong mode collapse. Nonetheless, failure cases do still occur. For example, on the first row of Fig.~\ref{fig:results}b we observed an undesired non-smooth transition from the fourth to last to the third to last frames. We also noticed that the temporal dynamics of the videos shown on the second and third rows of Fig.~\ref{fig:results}b is very similar, even though a frame-wise comparison shows that the videos are not the same. We refer to this effect as partial mode-collapse and believe this could be mitigated by increasing the number of discriminators when training $G_V$; this is left for future study. 
 
\begin{figure*}
\centering
\includegraphics[width=0.95\textwidth, height = 1.5cm]{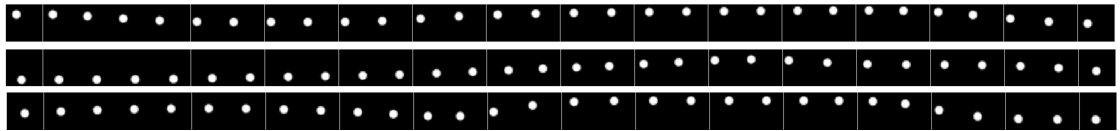}
\caption{Video generator samples for $G_F$ trained with one bouncing ball and $G_V$ trained with three bouncing balls.}
\label{fig:replace}
\end{figure*}

Moreover, time coherence was also studied in a case where $G_F$ and $G_V$ were trained using different datasets, namely bouncing balls with 1 and 3 balls, respectively. Notice that both one and three bouncing balls datasets have similar temporal dynamics and no further training to fine-tune $G_V$ after replacing $G_F$ was executed. Three samples from the new video generator are shown in Fig.~\ref{fig:replace}, from which one can notice that, even though the dynamics is not perfectly preserved in all frame transitions, as in some cases the ball changes its trajectory without hitting a wall first, smooth transitions between frames are still maintained. This simple experiment indicates that the video generator is indeed able to, at some extent, independently learn the temporal dynamics without specifically focusing on the content of each frame. 

Finally, we objectively assessed smoothness by measuring the mean-squared error (MSE) between consecutive frames. Average and standard deviation for 30 random samples drawn from models obtained using 3 and 1 bouncing balls are presented in Table \ref{tab:mse}. The same metrics are provided for real data and videos obtained using a random sequence of latent variables for comparison. For the 1-ball case, generated samples are as smooth as real videos. For the 3 ball cases, in turn, aforementioned eventual non-smooth transitions happen for sequences as long as 30 frames, as confirmed by the higher MSE, which is still much lower than random sequences.

\subsection{Investigating what the frame generator is learning}

\begin{figure}
\centering
\includegraphics[width=0.45\textwidth]{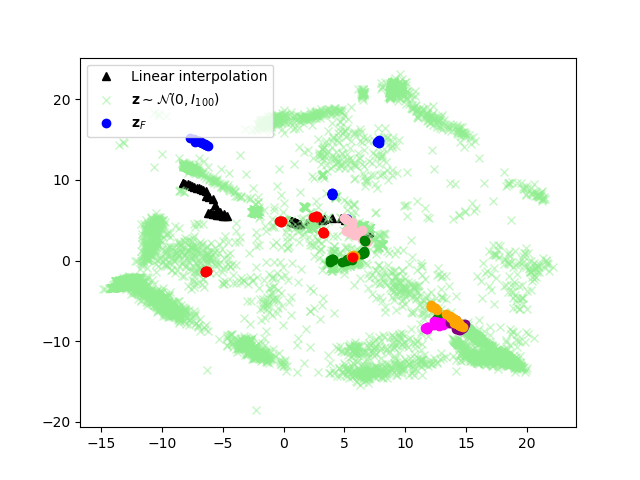}
\caption{Two-dimensional isomap obtained by plotting $z_{F_i}$ for 6 generated videos (circles, different colors represent different videos), samples from the prior (green crosses), and linear interpolations (black triangles).}
\label{fig:isomap} 
\end{figure}

We plot the isomap of the sequence of latent variables for 6 videos in order to investigate what the frames generator is learning. We included in this plot samples randomly drawn from the prior $\mathcal{N}(0, I_{100})$ with the aim of verifying whether $G_V$ is simply learning how to sample from the prior without any further knowledge. Another hypothesis we wanted to investigate is whether $G_V$ is learning to linearly interpolate latent variables. For that, we plot in the isomap two sequences of latent variables obtained by linearly interpolating two random samples from the prior. Results are shown in Fig.~\ref{fig:isomap}.

\begin{table}[] 
\centering
\caption{MSE between consecutive frames.}
\resizebox{0.8\columnwidth}{!}{
\begin{tabular}{cccc}
\hline
                         &                      & Mean & Std. dev.    \\ \hline
\multirow{3}{*}{3 balls} & Real data            & 0.0222  & 0.0005 \\
                         & Proposed             & 0.0735  & 0.0057 \\
                         & $z_{F_{i}} \sim \mathcal{N}(0, I_{100})$      & 0.2060  & 0.0061 \\ \hline
\multirow{3}{*}{1 ball} & Real data             & 0.0222  & 0.0005 \\
                         & Proposed             & 0.0227  & 0.0051 \\
                         & $z_{F_{i}} \sim \mathcal{N}(0, I_{100})$      & 0.0766  & 0.0023 \\ \hline
\end{tabular}}
\label{tab:mse}
\end{table}

By observing the obtained plot, we notice that samples from the prior (green crosses) are spread across the plane, while linear interpolations (black triangles) are concentrated in particular regions of the plane. The set of latent variables obtained with $G_V$ (circles, different colors represent different videos), on the other hand, seem to have a different behavior. In some cases, small clusters of $z_{F_i}$'s belonging to the same video are located in different parts of the isomap, which lead us to conclude that the video generator learns to ``jump'' across the manifold defined by $G_F$ whenever it is necessary.    


\section{Conclusions and future directions} \label{sec:conc}

We introduced a novel approach for unsupervised generation of temporal data using GANs. The method aims to break the problem into frame and sequence generation, and to solve them separately, thus making both tasks easier. Evaluation is performed on unsupervised video generation, and generated video samples presented good quality and diversity per frame as well as temporal coherence. This approach further provides indications regarding the structure of the implicit manifold parametrized by GANs, something that still remains elusive in the literature. Visualization of latent variables after dimensionality reduction via isomap indicates that the videos manifold is not continuous, as latent representations corresponding to visually similar frames are not necessarily close in the isomap. This work opens directions of future research as a general scheme for generative modeling of time-series. As such, we intend to apply the same approach to different domains. Further exploiting the video generation setting and including other objective video quality metrics is another target of future investigation.


\bibliography{bibliography.bib}
\bibliographystyle{IEEEbib.bst}

\end{document}